# Classification of histopathological breast cancer images using iterative VMD aided Zernike moments & textural signatures

Subhankar Chattoraj[$], Karan Vishwakarma[$]
[$]Techno India University, West Bengal

*Abstract—*. **In this paper we present a novel method for an automated diagnosis of breast carcinoma through multilevel iterative variational mode decomposition (VMD) and textural features encompassing Zernaike moments, fractal dimension and entropy features namely, Kapoor entropy, Renyi entropy, Yager entropy features are extracted from VMD components. The proposed method considers the histopathological image as a set of multidimensional spatially-evolving signals. ReliefF algorithm is used to select the discriminatory features and statistically most significant features are fed to squares support vector machine (SVM) for classification. We evaluate the efficiency of the proposed methodology on publicly available Breakhis dataset containing 7,909 breast cancer histological images, collected from 82 patients, of both benign and malignant cases. Experimental results shows the efficacy of the proposed method in outperforming the state of the art while achieving an average classification rates of 89.61% and 88:23% using three-fold and ten-fold cross-validation strategies, respectively. This system can aid the pathologist in accurate and reliable diagnosis of biopsy samples. BreaKHis, a publicly dataset available at http://web.inf.ufpr.br/vri/breast-cancer-database**

*Index Terms—* Medical imaging; Histopathological image classification; Breast cancer; Variational mode decomposition; LS-SVM; Zernike moments.

## I. INTRODUCTION

Cancer is regarded as a massive health problem in the world. It is estimated by the International Agency for Research on Cancer (IARC) that by 2012 there was 8.2 million deaths due to cancer part of world health organization (WHO) and by 2030 there would be 27 million new cases of cancer [1]. Excluding the skin cancer breast cancer (BC) in women is regarded as the most common type of cancer. In recent times, there has been lot of advancement for analyzing of BC progression and molecular markers related discovery but for the BC diagnosis histopathological analysis still remains the most widely used [2]. Despite of development in various diagnostic imaging technologies but the final BC diagnosis of the samples is done by the pathologist under microscope. Therefore a Computer-Aided Detection/Diagnosis (CAD/CADx) system is needed so that it can assist the pathologist to be more accurate and productive. In an image analysis system the classification of histopathology images into distinct histopathology patterns, corresponding to the non-cancerous or cancerous condition of the analyzed tissue, is often the primordial goal but the main challenge is the complexity involving in dealing with such histopathological images. For more than 40 years, automatic imaging process for cancer diagnosis purpose has been studied and evaluated as a research topic [3] but complexity still looms in image analysis. As an example in recent research [4] for nuclei segmentation divergent algorithm were compared and tested where the cases were stratify on dataset of 500 images of being malignant or benign and the accuracy percentage reported ranging from 96% to 100%. Another recent research [5] on diagnosis of BC was presented to distinguish the images on either being a benign or malignant based on cytological images analysis of needle biopsies. The researcher reports a performance accuracy of 98% on the 737 images implementing four unlike classifiers trained with a 25-dimensional feature vector. In similar to the study illustrated above a study [6] implementing support vector machine and neural network hinged on the nuclei segmentation of cytological images proposed a diagnosis system for BC. In this study the accuracy reported was 76% to 94%. In recent study cascade was proposed approach with an option of rejection. The easy cases are solved in the first level of the cascade, whereas the hard cases are dealt by the second level which has very complex pattern classification system. They evaluated the proposed method on the database which was proposed by the Israel Institute of Technology. The database was consisted of 361 images and reported the results of about 97% of reliability. In another work [8] done by the same authors, a recognition rate of about 92% was achieved when they evaluated an ensemble of one-class classifiers on the similar database. The majority of the recent works related to the BC classification are mainly focused on the Whole Slide Imaging (WSI) [7-9] [6] [4]. Nonetheless, there are many barriers to the wide adoption of WSI and the other forms of the digital pathology. These barriers include insufficient productivity for high-volume clinical routines, the huge cost of implementing and operating the technology, intrinsic technology-related concerns, unsolved regulatory issues, as well as "cultural resistance" from the pathologists [10]. Until recently, small datasets were used for most of the work based on the BC histopathology image analysis. These datasets were not usually available for the scientific community. In order to check this gap, a dataset composed of 7,909 breast histopathological images was introduced [11]. This dataset was acquired from 82 patients. In a similar study, done by the authors they evaluated different classifiers and six different textural descriptors, they reported a series of experiments with the accuracy rates ranging from 80% to 85%, depending on the image magnification factor. According to the results presented in the above study [11], it is certain that the texture descriptors may provide a good representation to train classifiers. Nevertheless, some researchers have supported the idea the main weakness of the current machine learning methods lies particularly on this feature engineering step [12-13]. For them, the machine learning algorithms must be able to extract and organize the discriminative information from the data and must be less dependent on feature engineering. In the other words, the machine learning algorithms must be capable of learning the representation. This learning is not a current method due to the popularization of Graphic Processing Units (GPUs) it has emerged as a feasible alternative for remitting computational high output at lower cost acquired through enormous parallel architecture. Convolutional Neural Network (CNN) [14] has been widely implemented to procure result as the revolutionary technique in pattern recognition [15-16]. In a recent study [17] in similar image classification of microscopic and macroscopic texture that CNN is able to transcend conventional textural descriptors. Traditional approach in feature extraction necessitates substantial effort and effectual expert knowledge in similar domain, regularly leading to customized solution for particular problem [18]. In this work, deep learning has been evaluated for BC histopathological image classification. To deal with high-resolution texture images architecture utilized for low-resolution images we have scrutinized different CNN architecture.

## II. PROPOSED METHODOLOGY

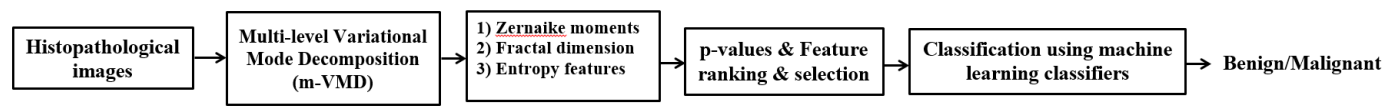

Fig. 1. Pipeline for VMD aided Zernike moments & textural signatures based classification.

### A. Dataset description

The microscopic biopsy images of benign and malignant breast tumors are included in the BreakHis database [11]. These images were gathered through a clinical study from January to December (2014), where all the patients referred to the P&D Lab, Brazil, with a clinical indication of BC were requested to take part in the study. The study was approved by the institutional review board and the written informed agreement was given by the patients participating in the study. The anonymity was maintained for all of the data. The samples are produced from the breast tissue biopsy slides which are stained with the hematoxylin and eosin (HE). The collection of samples is done by the surgical (open) biopsy (SOB), which are then prepared for the histological study and are designated by the pathologists of the P&D Lab. In this work, the standard paraffin process (which is extensively used in clinical routine) is being used as a preparation procedure. The main objective is to preserve the original tissue structure and the molecular composition so that it can be observed under the light microscope. The steps involved in the complete preparation procedure are the fixation, dehydration, clearing, infiltration, embedding, and trimming [20]. In order to mount the tissue sections on slides, sections of 3μm are cut with the help of using a microtome. Then after the staining, a glass coverslip is used to cover all the sections. After that the tumoral areas in each slide is identified by the anatomopathologists by visually analyzing the tissue sections under the microscope. In every case, the final diagnosis is produced by the experienced pathologists and is confirmed by the complementary exams such as the immunohistochemistry (IHC) analysis.

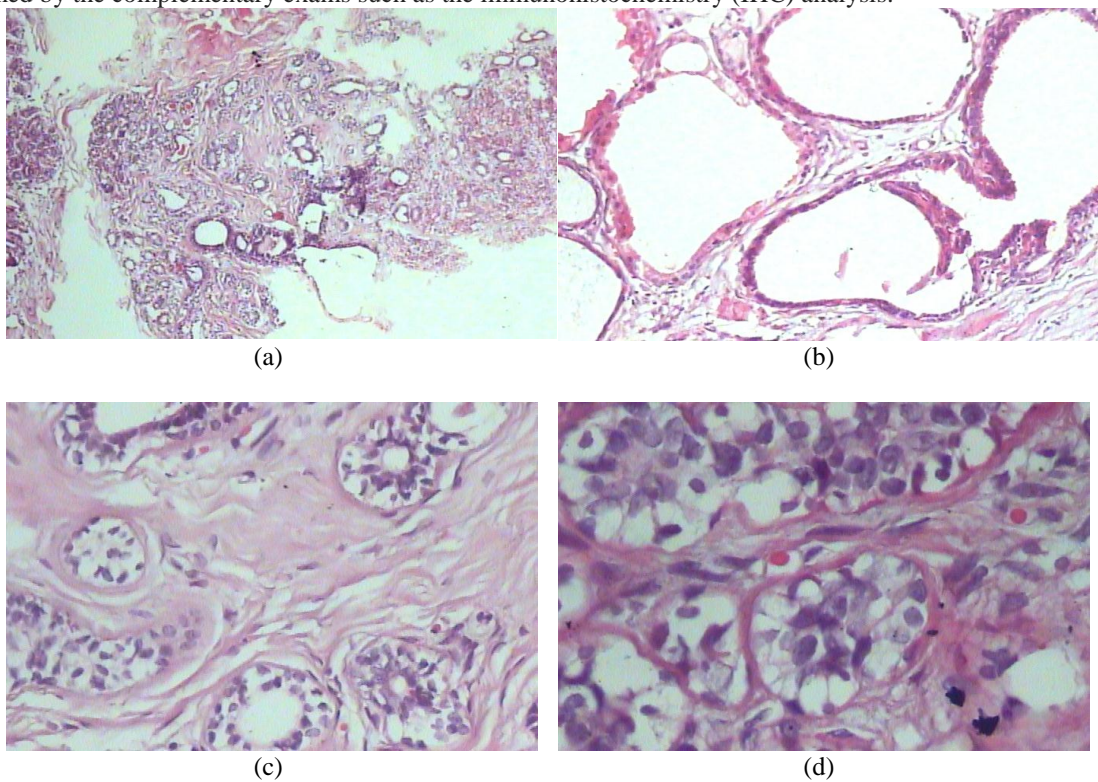

Fig. 2. A slide of breast malignant tumor (stained with HE) seen in different magnification factors: (a) 40X, (b) 100X, (c) 200X, and (d) 400X. Highlighted rectangle (manually added for illustrative purposes only) is the area of interest selected by pathologist to be detailed in the next higher magnification factor.

[Type here]

Table1: Image arrangement by magnification factor and class

| MAGNIFICATION | 40X | 100X | 200X | 400X | Total | Patient aggregate |
|---|---|---|---|---|---|---|
| MALIGNANT | 1370 | 1437 | 1390 | 1232 | 5429 | 58 |
| BENIGN | 625 | 644 | 623 | 588 | 2480 | 24 |
| TOTAL | 1995 | 2081 | 2013 | 1820 | 7909 | 82 |

*B. Variational mode decomposition (VMD)*

The underlying information is effectively captured with the help of signal processing techniques despite of the non-stationary nature of the signal. Unlike the EMD [21], the VMD [22] is a non-recursive method. The decomposition which is based on EMD highly relies upon the stopping criteria, interpolation and the extreme point finding. This deficiency of mathematical theory and degrees of freedom supports the idea of having a robust decomposition method. The VMD which is an alternative to the EMD method [22] is far more improved in comparison to the EMD and robust to noise. VMD relies on the frequency information content of the signal. It is an adaptive and non-stationary [22]. It has sparsity properties and the ability to decompose the signal that centers around particular frequency into the bandlimited modes. The multi-component signal can be decomposed non-recursively into the discrete number of bandlimited sub-signals through VMD.

1. For each sub-band modify the signal into its analytic equivalent.
2. Exponential term which is modified to specific calculated center frequency has been utilized to transpose frequency spectrum of individual sub-band
3. In order to estimate the bandwidth the squared L2-norm of the gradient is implemented.

For VMD, the constrained variational problem for a multi-component signal $n(t)$ can be illustrated as follows [22]:

$$\min_{p_k \omega_k} \left\{ \sum_k \left\| \partial t \left[ \left( \delta(t) + \frac{i}{\pi t} \right) *_{p_k}(t) \right] exp^{-i\omega_k t} \right\|_2^2 \right\} S.T. \sum_k p_k = n(t) \quad (1)$$

Where, $k^{th}$ decomposed bandlimited VMD component signifies $p_k$ and center frequency is signified by $\omega_k$. The constrained problem in equation (1) can be described by integrating the Lagrangian multiplier $\mu$ and the quadratic penalty. This can be rewritten as follows [22]:

$$\mathcal{L}(p_k, \omega_k, \mu) = a \sum_k \left\| \partial_t \left[ \left( \partial(t) + \frac{i}{\pi t} \right) *_{p_k}(t) \right] exp^{-i\omega_k t} \right\|_2^2 + \left\| n(t) - \sum_k p_k(t) \right\|_2^2 + \langle \mu(t), n(t) - \sum_k u_k(t) \rangle \quad (2)$$

As alternate direction method of multipliers (ADMM) in respect to equation (1) the saddle point of equation (2) can be derived [22]. $K^{th}$-mode estimation is refurbished as follows.

$$\hat{p}_k^{n+1}(\omega) = \frac{\hat{n}(\omega) - \sum_{j \neq k} \hat{p}_j(\omega) + \frac{\hat{\mu}(\omega)}{2}}{1 + 2a(\omega - \omega_k)^2} \quad (3)$$

Where, balancing parameter of the data fidelity constraint is denoted by $a$. The center frequency is upgraded as center of gravity, which can be presented as [22]:

$$\omega_k^{n+1} = \int_0^\infty \omega |\hat{p}_k(\omega)|^2 d\omega \bigg/ \int_0^\infty |\hat{p}_k(\omega)|^2 d\omega \quad (4)$$

In this study, we have employed the two dimensional VMD (2D-VMD) [23] in order to achieve the image decomposition. The 2D-VMD is applied in a repetitious manner on the green channel of the digital color fundus image. Repeatedly applying VMD can be very helpful in extracting the important details which can be used for the diagnosis of glaucoma. VMD components are

[Type here]

usually bandlimited. In order to capture the additional details of pixel variation, further decomposition of the previous VMD component is required. The fine details can be extracted from the previously decomposed VMD component with the help of the repetition process, which does the task of getting fine details by reducing the frequency band of the newly decomposed VMD component. The block diagram of the repetitive 2D-VMD is depicted in Fig. 3. In the first repetition, the image is decomposed into two modes. During the second repetition, the high-frequency band of the first repetition (2nd mode of the first repetition) is further decomposed into two modes. Fig. 3 depicts the decomposition of the 2nd mode of each repetition, where both of the 2nd modes of each repetition are decomposed again three more times. Therefore, we have acquired 10 VMD components with two components in each repetition.

*C. Feature extraction*

For apprehending relevant information from signals feature extraction plays a paramount role. Textural features are significant for automated glaucoma diagnosis. In image coarseness, pixel regularity and smoothness is computed by textural features [24]. In order to ensnare the pixel intensity variance in the image entropy and fractal dimension (FD) features have been extracted. We have implemented the entropy features namely Kapur entropy (KE) [25], Yager entropy (YE) [26] and Renyi entropy (RE) [25] in this proposed work. The measure of randomness and uncertainty is determined as entropy [27]. Distribution of Pixel intensity equally results to no information and therefore zeros entropy. Non-Shannon entropies are utilized in this work as the dynamic range is higher and makes preferable approximation of regularity and dispersion [26]. Let $q_m$ determines the probability pixel value $m$ happened $n$ times and let the size of the image be $c \times d$. Now $q_m$ can be expressed as $q_m = n/cXd$. Now RE, YE, and KE can be expressed as follows [26]:

$$YE = 1 - \frac{\sum_{m=0}^{X-1}|2q_m - 1|}{|cXd|} \quad (5)$$

$$RE = \frac{1}{1-a} \log_2 \left( \sum_{m=0}^{X-1} q_m^a \right) \quad (6)$$

$$KE = \frac{1}{b-a} \log_2 \left( \frac{\sum_{m=0}^{X-1} q_m^a}{\sum_{m=0}^{X-1} q_m^b} \right) \quad (7)$$

*D. Zernike moments (ZM)*

The Zernike moment (ZMs) [28] is described as orthogonal moments procured by forecasting the input image onto complex orthogonal Zernike polynomials. ZMs of order p and repetition q for a function *f(x, y)* over a unit disk are illustrated by:

$$Z_{pq} = \frac{p+1}{\pi} \iint_{x^2+y^2 \leq 1} f(x,y) V_{pq}^*(x,y) dx dy \quad (8)$$

where p defines non-negative integer, q defined as an integer such that $0 \leq |q| \leq p$ and $p - |q| = even$. Complex conjugate of the Zernike orthogonal basis function defines $V_{pq}^*(x, y)$. Now $V_q(x, y)$ is defined by:

$$V_{pq}(x, y) = V_{pq}(r, \theta) = R_{pq}(r) \exp(jp\theta) \quad (9)$$

where $= \sqrt{x^2 + y^2}$, $\theta = \tan^{-1}(x/y), 0 \leq \theta \leq 2\pi, j = \sqrt{-1}$. The radial polynomial $R_{pq}(r)$ is defined as:

$$R_{pq}(r) = \sum_{s=0}^{(p-|q|)/2} (-1)^2 \frac{(p-s)!}{s!\left(\frac{p+|q|}{2} - s\right)!\left(\frac{p-|q|}{2} - s\right)!} (r)^{p-2s} \quad (10)$$

In Equation (9) the zeroth order approximation is [29-30]:

[Type here]

$$Z_{pq} = \frac{2(p+1)}{\pi N^2} \sum_{i=0}^{N-1} \sum_{k=0}^{N-1} f(x_i y_k) R_{pq}(r_{ik}) e^{-jq\theta_{ik}} \qquad (11)$$

where the normalized coordinate location is defined by $(x_i y_k)$ which is commensurate to pixel $(i,k)$ which is derived by implementing coordinate transformation given in following:

$$x_i = \frac{2i+1-N}{N\sqrt{2}}, y_k = \frac{2k+1-N}{N\sqrt{2}} \qquad (12)$$

for all i, k = 0, 1, 2, . . ., N − 1. For ZMs computation we utilize other unit disk which provides superior performance for ZMs-based image pattern matching complication [29-39]. For ZMs-based approach as proposed in your system, we are not using moments with negative q, as $|Z_{p,q}| = |Z_{p,-q}|$.

*E. Fractal dimension (FD)*

Fractals are the objects which comprises of self-similarity and irregularity [31]. FD is an estimation of irregularity, self-similarity and roughness. As it comprises of scale dependent property it is very convenient for studying texture [31]. Malignant and benign images have textural difference, which can be apprehended using FD. The value of FD increases with the higher roughness of surface or texture [24]. If a surface $S$, ascend up or down by a factor $f$, then it is self-similar only if $S$ is union of non-overlapping copies ($S_f$) of itself when extended. This similarity can be computed utilizing FD [31]:

$$1 = S_f f^{FD} \qquad (13)$$

This can also be rewritten as follows:

$$FD = \frac{\log(S_f)}{\log\left(\frac{1}{f}\right)} \qquad (14)$$

where, $f = scaled\ value/original\ value$

To determine the FD, we have implemented Differential box counting (DBC) [32]. The DBC has been modified and named sequential box counting (SBC). Initial size of the grid in the SBC method is set to 2 and final one is set to the image size $c\ X\ d$. The $S_f$ in (13) is sum of differences of maximum and minimum intensities of each 2 X 2 grid. This process is repeated until the grid is equal to image size.

*F. LS-SVM & kernel selection*

In the computer-based diagnosis application classifier plays a crucial role. For classification of various classes a decision boundary is created utilizing hyper-planes. LS-SVM [36, 37] has been implemented for classification of diverse biomedical signals such as heart rate [38], heart sound [39] and electroencephalogram [38]. In order to separate non-linear features using LS-SVM, radial basis function (RBF) [40] as a non-linear kernel function is implemented that maps the input space into a higher dimensional space so that it can be linearly separable. LS-SVM is a oversee machine learning algorithm in compute-based glaucoma diagnosis [34].

*III. RESULTS AND DISCUSSION*

In this section, details of our experiment, results and key finding are discussed. In our experiment we have utilized the breaKHis dataset which is available online and the illustration of the dataset has been given in Table1. We have randomly utilized 58 patients (70%) for training and the rest 25 patients (30%) for testing. The classifier which is mentioned earlier has been trained using the images of 58 patients and also 5 try-out of arbitrary selection of testing data. The images of 25 patients have been utilized to test the training model. In order to compute our results the evaluation metrics used are discussed in the following section.   .

Evaluation metric
We have utilized the patient recognition rate (PRR) as evaluation metric to contrast the results available in the existing approach. PRR denotation is given below:

[Type here]

$$PRR = \frac{\sum_{i=1}^{N} PS_i}{N}$$

Where N defines the number of patients in total (accessible for testing). The score of patients is given below.

$$PS = \frac{N_{rec}}{N_P}$$

Where $N_{rec}$ are correctly classify and $N_P$ is the count of cancer images in total.

TABLE VIII – OVERALL PERFORMANCE ANALYSIS WITH PRE-PROCESSED IMAGES

| Zoom factor | Performance Metrics | | | | |
| --- | --- | --- | --- | --- | --- |
| | Accuracy (Acc.) (%) | Sensitivity (Sen.) (%) | Specificity (Spec.) (%) | Positive Predictive Value (PPV) (%) | Negative Predictive Value (NPV) (%) |
| 40X | 87.7 | 88.4 | 85.3 | 84.6 | 88.6 |
| 100X | 85.8 | 87.1 | 85.1 | 84.6 | 87.5 |
| 200X | 88.0 | 85.7 | 87.6 | 83.4 | 85.1 |
| 400X | 84.6 | 83.7 | 83.5 | 85.4 | 84.3 |
| Full Dataset | 87.0 | 86.9 | 85.3 | 84.7 | 88.4 |

TABLE-I: A COMPARISON OF SELECTED STUDIES IN THE AUTOMATED CLASSIFICATION BREAST MALIGNANT HISTOPATHOLOGICAL IMAGES USING THE BREAKHIS DATABASE

| Author, Years | Feature descriptor | Classifier | Performance | | | |
| --- | --- | --- | --- | --- | --- | --- |
| | | | 40x | 100x | 200x | 400x |
| Fabio A. Spanhol., et al, 2016 [1] | LBP | 1-NN, QDA, RF, SVM | 75.6 | 73.2 | 72.9 | 73.1 |
| | CLBP | 1-NN, QDA, RF, SVM | 77.4 | 76.4 | 70.2 | 72.8 |
| | LPQ | 1-NN, QDA, RF, SVM | 73.8 | 72.8 | 74.3 | 73.7 |
| | GLCM | 1-NN, QDA, RF, SVM | 74.7 | 78.6 | 83.4 | 81.7 |
| | ORB | 1-NN, QDA, RF, SVM | 74.4 | 69.4 | 69.6 | 67.6 |
| | PFTAS | 1-NN, QDA, RF, SVM | 83.8 | 82.1 | 85.1 | 82.3 |
| Fabio A. Spanhol, et al, 2016 [2] | Deep features | CNN | 84.0 | 83.9 | 86.3 | 82.1 |
| Bayramoglu., et.al, 2016 [3] | Deep Features | Multi task CNN | 81.87 | 83.39 | 82.56 | 80.69 |
| Fabio A. Spanhol., et al, 2016 [4] | Deep Features | CNN & Fusion Rules | 90.0 | 88.4 | 84.6 | 86.1 |
| Kosmas Dimitriopoulos., et al, 2017 [5] | Grassmannian manifold | VLAD encoding | 91.8 | 92.2 | 91.6 | 90.5 |
| Zyongyi Han., et. al, 2017 [6] | Deep Features | Structured Deep Learning | 95.8 | 96.9 | 96.7 | 94.9 |
| **This work** | **Zernike moments** | **LS-SVM** | **87.7** | **85.8** | **88.0** | **84.6** |

## IV. CONCLUSION

In this paper, we have presented a set of experiments conducted on the BreaKHis dataset using a deep learning approach to avoid hand-crafted features. We have shown that we could use an existing SA architecture, in our case AlexNet that has been designed for classifying color images of objects and adapt it to the classification of BC histopathological images. We have also proposed several strategies for training the SA architecture, based on the extraction of patches obtained randomly or by a sliding window

[Type here]

mechanism, that allow to deal with the high-resolution of these textured images without changing the SA architecture designed for low-resolution images. Our experimental results obtained on the BreaKHis dataset showed improved accuracy obtained by SA when compared to traditional machine learning models trained on the same dataset but with state of the art texture descriptors. Future work can explore different SA architectures and the optimization of the hyper parameters. Also, strategies to select representative patches in order to improve the accuracy can be explored.

ACKNOWLEDGEMENT

The authors would like to thank Fabio Alexandre Spanhol for providing an exhaustive and inclusive discussion and standard baseline results of the BreaKHis dataset.

[Type here]